\documentclass[10pt,twocolumn,letterpaper]{article}

\usepackage{iccv}
\usepackage{times}
\usepackage{epsfig}
\usepackage{graphicx}
\usepackage{amsmath}
\usepackage{amssymb}
\usepackage{caption}
\usepackage{booktabs}
\usepackage{multirow}
\usepackage{ctable}

\usepackage{lib}
\usepackage{subcaption}
\usepackage{enumitem}
\usepackage[pagebackref=true,breaklinks=true,letterpaper=true,colorlinks,bookmarks=false]{hyperref}

\iccvfinalcopy 
\def\iccvPaperID{****} 

\begin{document}
\title{
Learning Unsupervised Multi-View Stereopsis
\\ via Robust Photometric Consistency
}

\author{First Author\\
Institution1\\
Institution1 address\\
{\tt\small firstauthor@i1.org}
\and
Second Author\\
Institution2\\
First line of institution2 address\\
{\tt\small secondauthor@i2.org}
}
\author{Tejas Khot$^*$ $^1$, Shubham Agrawal$^*$ $^1$, Shubham Tulsiani $^2$, Christoph Mertz $^1$, Simon Lucey $^1$, Martial Hebert $^1$\\
$^1$Carnegie Mellon University, $^2$ Facebook AI Research\\
{\tt\small $^1$\{tkhot, sagrawa1, cmertz, slucey, mhebert\}@andrew.cmu.edu}, \tt\small $^2$shubhtuls@fb.com}

\newcommand\blfootnote[1]{%
  \begingroup
  \renewcommand\thefootnote{}\footnote{#1}%
  \addtocounter{footnote}{-1}%
  \endgroup
}

\twocolumn[{%
\renewcommand\twocolumn[1][]{#1}%
\vspace{-1em}
\maketitle
\vspace{-1em}
\begin{center}
   \centering \includegraphics[width=\textwidth]{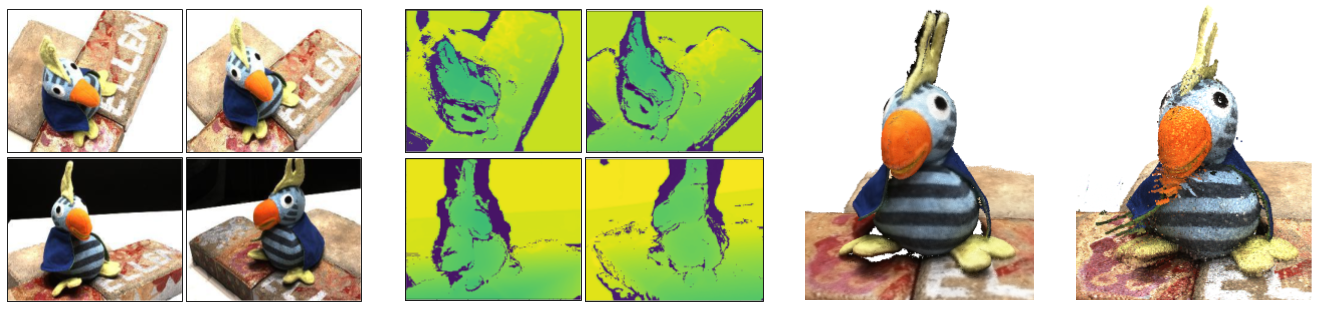} \captionof{figure}{Our model consumes a collection of calibrated images of a scene from multiple views and produces depth maps for every such view. We show that this depth prediction model can be trained in an unsupervised manner using our robust photo consistency loss. The predicted depth maps are then fused together into a consistent 3D reconstruction which closely resembles and often improves upon the sensor scanned model. Left to Right: Input images, predicted depth maps, our fused 3D reconstruction, ground truth 3D scan.}
   \figlabel{cover}
\end{center}%
}]


\begin{abstract}
\vspace{-3mm}
We present a learning based approach for multi-view stereopsis (MVS). While current deep MVS methods achieve impressive results, they crucially rely on ground-truth 3D training data, and acquisition of such precise 3D geometry for supervision is a major hurdle. Our framework instead leverages photometric consistency between multiple views as supervisory signal for learning depth prediction in a wide baseline MVS setup.  However, naively applying photo consistency constraints is undesirable due to occlusion and lighting changes across views. To overcome this, we propose a robust loss formulation that: a) enforces first order consistency and b) for each point, selectively enforces consistency with some views, thus implicitly handling occlusions. We demonstrate our ability to learn MVS without 3D supervision using a real dataset,
and show that each component of our proposed robust loss
results in a significant improvement.
We qualitatively observe that our reconstructions are often more complete than the acquired ground truth, further showing the merits of this approach.
Lastly, our learned model generalizes to novel settings, and our approach allows adaptation of existing CNNs to datasets without ground-truth 3D by unsupervised finetuning.
Project webpage:
\href{https://tejaskhot.github.io/unsup\_mvs}{https://tejaskhot.github.io/unsup\_mvs}.


\blfootnote{$^*$ The first two authors procrastinated equally on this work.}
\end{abstract}

\section{Introduction}
Recovering the dense 3D structure of a scene from its images has been a long-standing goal in computer vision. Several approaches over the years have tackled this multi-view stereopsis (MVS) task by leveraging the  underlying geometric and photometric constraints -- a point in one image projects on to another along the epipolar line, and the correct match is photometrically consistent. While operationalizing this insight has led to remarkable successes, these purely geometry based methods reason about each scene independently, and are unable to implicitly capture and leverage generic priors about the world \eg surfaces tend to be flat, and therefore sometimes perform poorly when signal is sparse \eg textureless surfaces.

To overcome these limitations, an emergent line of work has focused on learning based solutions for the MVS task, typically training CNNs to extract and incorporate information across  views. While these methods yield impressive performance, they crucially rely on ground-truth 3D data during the learning phase. We argue that this form of supervision is too onerous, is not naturally available, and it is therefore of both practical and scientific interest to pursue solutions that do not rely on such 3D supervision.

We build upon these recent learning-based MVS approaches that present CNN architectures with geometric inductive biases, but with salient differences in the form of supervision used to train these CNNs. Instead of relying on ground-truth 3D supervision, we present a framework for learning multi-view stereopsis in an \emph{unsupervised} manner, relying only on a training dataset of multi-view images. Our insight that enables the use of this form of supervision is akin to the one used in classical methods -- that the correct geometry would yield photometrically consistent reprojections, and we can therefore train our CNN by minimizing the reprojection error.

While similar reprojection losses have been successfully used by recent approaches for other tasks \eg monocular depth estimation, we note that naively applying them for learning MVS is not sufficient. This is because different available images may capture different visible aspects of the scene. A particular point (pixel) therefore need not be photometrically consistent with all other views, but rather only those where it is not occluded. Reasoning about occlusion explicitly to recover geometry, however, presents a chicken-and-egg problem, as estimates of occlusion depend on geometry and vice-versa. To circumvent this, we note that while a correct estimate of geometry need not imply photometric consistency with all views, it should imply consistency with at least \emph{some} views. Further, the lighting changes across views in an MVS setup are also significant, thereby making enforcing consistency only in pixel space undesirable, and our insight is to additionally enforce gradient-based consistency. 
We present a robust reprojection loss that enables us to capture these two insights, and allow learning MVS with the desired form of supervision.
Our simple, intuitive formulation allows handling occlusions without ever explicitly modeling them. 
Our setup and sample outputs are depicted in \figref{cover}. Our model, trained without 3D supervision, takes a collection of images as input and predicts per-image depth maps, which are then combined to obtain a dense 3D model.






\vspace{1mm}
\noindent In summary, our key contributions are:
\begin{itemize}
    \item A framework to learn multi-view stereopsis in an unsupervised manner, using only images from novel views as supervisory signal.
    \item A robust multi-view photometric consistency loss for learning unsupervised depth prediction that allows implicitly overcoming lighting changes and occlusion across training views.
\end{itemize}

\section{Related Work}

\noindent \textbf{Multi-view Stereo Reconstruction.} 
There is a long and rich history of work on MVS. We only discuss representative works here and refer the interested readers to \cite{seitz2006comparison,furukawa2015multi} for excellent surveys. There are four main stages in an MVS pipeline: view selection, propagation scheme, patch matching and depth map fusion. Schemes for aggregating multiple views for each pixel have been studied in \cite{kang2001handling,galliani2015massively,zheng2014patchmatch,goesele2007multi,schonberger2016pixelwise,kang2001handling}, and our formulation can be seen as integrating some of these ideas via a loss function during training.
The seminal work of PatchMatch\cite{barnes2009patchmatch} based stereo matching replaced the classical seed-and-expand\cite{furukawa2010accurate,goesele2007multi} propagation schemes. PatchMatch has since been used for multi-view stereo\cite{zheng2014patchmatch,galliani2015massively,schonberger2016pixelwise} in combination with iterative evidence propagation schemes, estimation of depth and normals. Depth map fusion\cite{shen2013accurate,schonberger2016pixelwise,jancosek2011multi,hu2012least,zach2008fast} combines individual depth maps into a single point cloud while ensuring the resulting points are consistent among multiple views and incorrect estimates are removed. Depth representations continue to dominate MVS benchmarks~\cite{aanaes2016large,schops2017multi} and methods seeking depth images as output thus decouple the MVS problem into more tractable pieces.

\vspace{2mm}
\noindent \textbf{Learning based MVS.}
The robustness of features learned using CNNs makes them a natural fit for the third step of MVS: matching image patches. CNN features have been used for stereo matching~\cite{hartmann2017learned,zbontar2016stereo} while simultaneously using metric learning to define the notion of similarity~\cite{Han2015MatchNetUF}. These approaches require a series of post-processing steps~\cite{hirschmuller2008stereo} to finally produce pairwise disparity maps. There are relatively fewer works that focus on learning all steps of the MVS pipeline. Volumetric representations encode surface visibility from different views naturally which has been demonstrated in~ \cite{ji2017surfacenet,kar2017learning}. These methods suffer from the common drawbacks of this choice of representation making it unclear how they can be scaled to more diverse and large-scale scenes. In ~\cite{kendall2017end}, a cost volume is created using CNN features and disparity values are obtained by regression using a differentiable soft argmin operation. Combining the merits of above methods and borrowing insights from classical approaches, recent works~\cite{mvsnet,huang2018deepmvs,wang2018mvdepthnet} produce depth images for multiple views and fuse them to obtain a 3D reconstruction. Crucially, all of the above methods have relied on access to 3D supervision and our work relaxes this requirement.

\vspace{2mm}
\noindent \textbf{Unsupervised depth estimation.}
With a similar motivation of reducing the requirement of supervision, several recent  monocular~\cite{godard2017unsupervised,Garg2016UnsupervisedCF,Kuznietsov2017SemiSupervisedDL} or binocular stereo based~\cite{Zhong2017SelfSupervisedLF} depth prediction methods have leveraged photometric consistency losses. As supervision signal, these rely on images from stereo pairs~\cite{godard2017unsupervised,Garg2016UnsupervisedCF,Kuznietsov2017SemiSupervisedDL} or monocular videos \cite{Xie2016Deep3DFA,Zhou2017UnsupervisedLO} during training. 
As means for visibility reasoning, the network is made to predict an explainability~\cite{Zhou2017UnsupervisedLO}, invalidation~\cite{Zhang2018ActiveStereoNetES} mask or by incorporating a probabilistic model of observation confidence~\cite{DBLP:conf/eccv/KlodtV18}. These methods operate on a narrow baseline setup with limited visual variations between frames used during training, and therefore do not suffer significantly due to occlusions and lighting changes. As we aim to leverage photometric losses for learning in an MVS setup, we require a robust formulation that can handle these challenges.




\section{Approach}
The goal in the MVS setup is to reconstruct the dense 3D structure of a scene given a set of input images, where the associated intrinsics and extrinsics for these views are known -- these parameters can typically be estimated via a preceding Structure-from-Motion (Sfm) step. While there are several formulations of the MVS problem focused on different 3D representations \cite{kar2017learning,furukawa2015multi,furukawa2010accurate}, we focus here on depth-based MVS setup. We therefore infer the per-pixel depth map associated with each input, and the dense 3D scene is then obtained via back-projecting these depth maps into a combined point cloud.

We leverage a learning based system for the step of predicting a depth map, and learn a CNN that takes as input an image with associated neighboring views, and predicts a per-pixel depth map for the central image. Unlike previous learning based MVS methods which also adopt a similar methodology, we only rely on the available multi-view images as supervisory signal, but do not require a ground-truth 3D scene. Towards leveraging this supervision, we build upon insights from classical methods, and note that the accurate geometry prediction for a point (image pixel) should yield photometrically consistent predictions when projected onto other views. We operationalize this insight and use a photometric consistency loss to train our depth prediction CNN, penalizing discrepancy between pixel intensities in original and available novel views. However, we note that the assumption of photometric consistency is not always true. The same point is not necessarily visible across all views. Additionally, lighting changes across views would lead to further discrepancy between pixel intensities. To account for possible lighting changes, we add a first-order consistency term in the photometric loss and therefore also ensure that gradients match in addition to intensities. We then implicitly deal with possible occlusions by proposing a robust photometric loss, which enforces that a point should be consistent with \emph{some}, but not necessarily all views.

We describe the architecture of the CNN used to infer depth in \secref{arch}, and present in \secref{photo} the vanilla version of photometric loss that can be used to learn this CNN in an unsupervised manner. We then present our robust photometric loss in \secref{robust}, and describe the overall learning setup, additional priors and implementation details in \secref{learnsetup}. While we primarily focus on the learning of the depth prediction CNN in this section, we briefly summarize how the learned CNN is integrated in a standard MVS setup at inference in \secref{inference}.

\subsection{Network architecture}
\seclabel{arch}
\begin{figure}
    \centering
    \includegraphics[width=\linewidth]{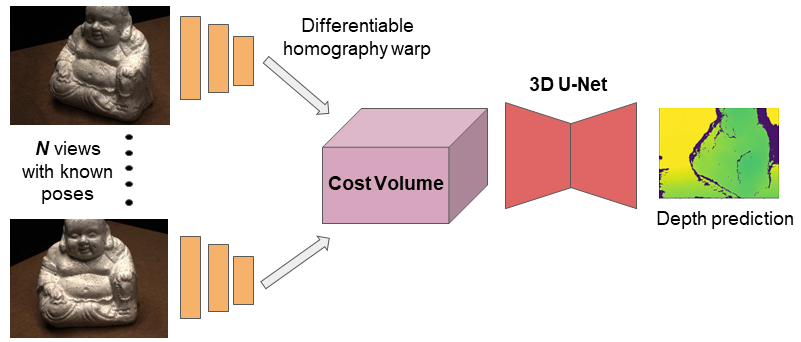}
    \caption{\textbf{Overview of our network.} We take as input $N$ images of a scene. Image features are generated using a CNN. Using differentiable homography, a cost volume is constructed by warping image features over a range of depth values. The cost volume is then refined using a 3D U-Net style CNN. The final output is a depth map at a downsized resolution. Details of the network architecture can be found in the supplemental.}
    \label{fig:network}
\end{figure}
The unsupervised learning framework we propose is agnostic to network architecture. 
Here, we adopt the model proposed in \cite{mvsnet} as a representative network architecture while noting that similar architectures have also been proposed in \cite{huang2018deepmvs,wang2018mvdepthnet}. 
The network takes as input $N$ images, extracts features using a CNN, creates a plane-sweep based cost volume and infers a depth map for every reference image.
A sketch of the architecture is given in Figure \ref{fig:network}. 
The emphasis of our work is on a way to train such CNNs in an unsupervised manner using a robust photometric loss, as described in the following sections.

\subsection{Learning via Photometric Consistency}
\seclabel{photo}

We now describe how our depth prediction network can be trained effectively without requiring ground truth depth maps. The central idea is to use a warping-based view synthesis loss, that has been quite effective in the stereo and monocular depth prediction tasks \cite{Zhou2017UnsupervisedLO,mahjourian2018unsupervised} though hasn't been explored for unstructured multi-view scenarios. Given an input image $I_s$, and additional neighboring views, our CNN outputs a depth map $D_s$. During training, we also have access to $M$ additional novel views of the same scene $\{I^m_v\}$, and use these to supervise the predicted depth $D_s$.


For a particular pair of views $(I_s, I^m_v)$ with associated intrinsic/relative extrinsic ($K, T$) parameters, the predicted depth map $D_s$ allows us to ``inverse-warp'' the novel view to the source frame using a spatial transformer network \cite{DBLP:journals/corr/JaderbergSZK15} followed by differentiable bilinear sampling to yield $\hat{I}_s^i$. 
For a pixel $u$ in the source image $I_s$, we can obtain its coordinate in the novel view with the warp:
\begin{equation}
    \hat{u} =   K~T(D_{s}(u)~\cdot K^{-1}~u)
\end{equation}

The warped image can then be obtained by bilinear sampling from the novel view image around the warped coordinates:
\begin{equation}
    \hat{I}_s^m(u) =  I^m_{v}(\hat{u})
\end{equation}

Alongside the warped image, a binary validity mask $V_s^m$ is also generated, indicating ``valid" pixels in the synthesized view as some pixels project outside the image boundaries in the novel view. As previously done in context of learning monocular depth estimation ~\cite{Zhou2017UnsupervisedLO}, we can then formulate a photo-consistency objective specifying that the warped image should match the source image. In our scenario of a multi-view system, this can naively be extended to an inverse-warping of all $M$ novel views to the reference view, with the loss being:
\begin{equation}\label{naive}
    L_{photo} = \sum_m^{M} || (I_s - \hat{I}_s^m) \odot V_s^m ||
\end{equation}

This loss allows us to learn a depth prediction CNN without ground-truth 3D, but there are several issues with this formulation \eg inability to account for occlusion and lighting changes. While similar re-projection losses have been successfully used in datasets like KITTI\cite{Geiger2012AreWR} for monocular or stereo reconstruction, there is minimal disocclusion and lighting change across views in these datasets.  However in MVS datasets, self-occlusion, reflection and shadows are a much bigger concern
. We therefore extend this photometric loss and propose a more robust formulation appropriate for our setup.



\subsection{Robust Photometric Consistency for MVS}
\seclabel{robust}
\begin{figure}
    \centering
    \includegraphics[scale=0.25]{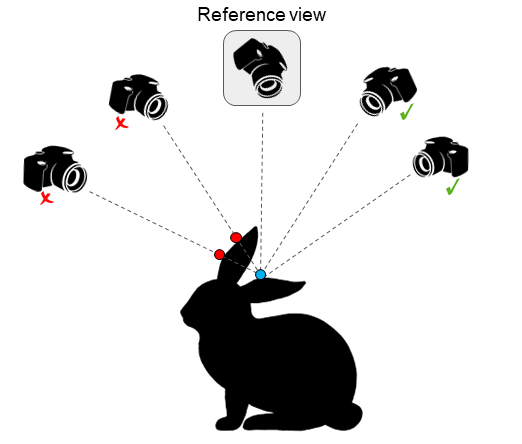}
    \caption{For a set of images of a scene, a given point in a source image may not be visible across all other views.}
    \label{fig:tempo}
\end{figure}

\begin{figure}
    \centering
    \includegraphics[width=\linewidth]{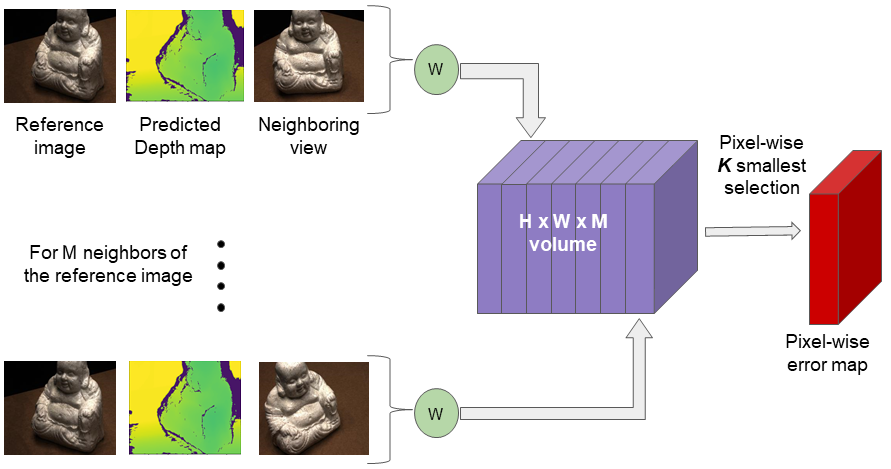}
    \caption{Visualization of the robust pixel-wise aggregation loss used for training. The predicted depth map from the network, along with the reference image are used to warp and calculate a loss map for each of $M$ non-reference neighboring views, as given in eqn \ref{gipuma}. These $M$ loss maps are then concatenated into a volume of dimension $H\times W \times M$, where $H$ and $W$ are the image dimensions. This volume is used to perform a pixel-wise selection to pick the $K$ ``best" (lowest loss) values, along the 3rd dimension of the volume (i.e. over the $M$ loss maps), using which we take the mean to compute our robust photometric loss.}
    \label{fig:loss}
\end{figure}



Our proposed robust photometric loss formulation is based on two simple observations -- image gradients are more invariant to lighting changes than intensities, and that a point need only be photometrically consistent with some (and not all) novel views.

The first modification we make in fact leverages  insights developed over many years of MVS research \cite{galliani2015massively}, where a number of conventional approaches have found that a matching cost based on both the absolute image intensity and the difference of image gradients works much better than just the former.  We also found that due to the large variations in pixel intensities between images, it is important to take a huber loss for the absolute image difference term. 
The inverse-warping based photometric loss of eqn \ref{naive} is therefore modified to reflect this: 
\begin{equation}\label{gipuma}
    L_{photo} = \sum_{m=1}^{M} ||  (I_s - \hat{I}_s) \odot V_{s}^m ||_{\epsilon} + || (\nabla I_s - \nabla \hat{I}_s^m) \odot V_{s}^m  ||
\end{equation}
We refer to this as a first-order consistency loss.

We next address the issues raised by occlusion of the 3D structure in the different images. The loss formulations discussed above enforce that each pixel in the source image should be photometrically consistent with \emph{all} other views. As shown in Fig~\ref{fig:tempo}, this is undesirable as a particular point may only be visible in a subset of novel views due to occlusion.
Our key insight is to enforce per-pixel photo-consistency with only top-$K$ (out of $M$) views. Let $L^m(u)$ denote the first-order consistency loss for a particular pixel $u$ w.r.t a novel view $I_m^i$. Our final robust photometric loss can be formulated as:
\begin{equation} \label{robust}
L_{photo} = \sum_{u} \min_{\substack{m_1, \cdots m_K \\ m_i \neq m_j \\ V_{s}^{m_k}(u) > 0 }} ~~ \sum_{m_k} L^{m_k}(u)
\end{equation}

The above equation simply states that for each pixel $u$, among the views where the pixel projection is valid, we compute a loss using the best $K$ disjoint views. An illustration of this is shown in Fig~\ref{fig:loss}. To implement this robust photometric loss, we inverse-warp the $M$ novel-view images to the reference image and compute a per-pixel first order consistency ``loss-map". All $M$ loss-maps are then stacked up into a 3D loss volume of dimensions $W \times H \times M$. For each pixel, we find the $K$ least value entries with valid mask, and sum them to obtain a pixel-level consistency loss.




\subsection{Learning Setup and Implementation Details}
\seclabel{learnsetup}

During training, the input to our depth prediction network comprises of a source image and $N=2$ additional views. However, we enforce the photometric consistency using a larger set of views ($M=6, K=3$). This allows us to extract supervisory signal from a larger set of images, while only requiring a smaller set at inference.




In addition to the robust photometric losses above, we add structured similarity ($L_{SSIM}$) and depth  smoothness ($L_{Smooth}$) objectives suggested by ~\cite{mahjourian2018unsupervised} for monocular depth prediction task. The smoothness loss enforces an edge-dependent smoothness prior on the predicted disparity maps. The SSIM loss is a higher order reconstruction loss on the warped images, but as it is based on larger image patches, we do not apply our pixel-wise selection approach for the robust photometric loss here. Instead, the two neighboring views with the highest view selection score are used to calculate SSIM loss. We describe the formulation in more detail in the appendix.

Our final end-to-end unsupervised learning objective is a weighted combination of the losses described previously: 
\begin{equation}
 L = \sum \alpha L_{photo} + \beta L_{SSIM} + \gamma L_{Smooth}
\end{equation}

For all our experiments, we use $\alpha = 0.8$ , $\beta = 0.2$ and $\gamma = 0.0067$. The network is trained with ADAM\cite{Kingma2014AdamAM} optimizer, learning rate of 0.001 and a 1st moment decay factor of 0.95. We use Tensorflow \cite{abadi2016tensorflow} to implement our learning pipeline.
As also noted by~\cite{mvsnet}, the high GPU memory requirements of the network imply that it is efficient to use a smaller image resolution and coarser depth steps at training, while a higher setting can be used for evaluation. We note the image resolutions used in the Experiments section.


\subsection{Inference using Learned Depth Prediction}
\seclabel{inference}

At test time, we take a set of images of a 3D scene, and predict the depth map of each image through our network. This is done by passing one reference image and 2 neighboring images through the network, which are chosen on the basis of the camera baselines or a view selection score if available.
The set of depth images are then fused to form the point cloud. We use Fusibile~\cite{galliani2015massively}, an open source utility, for the point cloud fusion.

\begin{table*}[ht]
\centering
\caption{Quantitative results on \textit{DTU}'s evaluation set \cite{aanaes2016large}. We evaluate two classical MVS methods (top), two learning based MVS methods (bottom) and three unsupervised methods (naive photometric baseline and two variants of our robust formulation) using both the distance metric \cite{aanaes2016large} (lower is better), and the percentage metric \cite{knapitsch2017tanks} (higher is better) with respectively $1mm$, $2mm$ and $3mm$ thresholds}
\begin{tabular}{c c c c | c c c | c c c | c c c}
\specialrule{.16em}{.08em}{.08em} 
          & \multicolumn{3}{c|}{Mean Distance (mm)}               & \multicolumn{3}{c|}{Percentage (\textless $1mm$)}  & \multicolumn{3}{c|}{Percentage (\textless $2mm$)} & \multicolumn{3}{c}{Percentage (\textless $3mm$)}\\ 
          & \multicolumn{3}{c|}{Acc. Comp.  \textit{overall}}     & \multicolumn{3}{c|}{Acc.           Comp.          \textit{f-score}}  & \multicolumn{3}{c|}{Acc.           Comp.           \textit{f-score}}  & \multicolumn{3}{c}{Acc.           Comp.           \textit{f-score}}\\ \hline
Furu \cite{furukawa2010accurate}       & 0.612          & 0.939          & 0.775          & 69.37          & 57.97  & 63.16          & 77.30          & 64.06          & 70.06  & 79.77 & 66.27 & 72.40       \\ 
Tola \cite{Tola2011EfficientLM}      & 0.343          & 1.190          & 0.766          & 88.96          & 53.88          & 67.12          & 92.35          & 60.01         & 72.75 & 93.46 & 62.29  & 74.76        \\ 
\hline
Photometric &  1.565          & 1.378 & 1.472 & 46.90          & 42.16  & 44.40 & 71.68          & 55.90 & 62.82 & 81.92 & 60.56 & 69.64\\
Ours (Photometric+G)    &  1.069          & 1.020 & 1.045 & 55.98          & 45.24  & 50.04 & 81.11          & 60.70 & 69.43 & 87.03 & 64.36 & 74.00\\
Ours (Robust: G + top-K)    &  0.881          & 1.073 & 0.977 & 61.54          & 44.98 & 51.98 & 85.15          & 61.08 & 71.13 & 89.47 & 64.26 & 74.80\\
\hline
SurfaceNet\cite{ji2017surfacenet} & 0.450          & 1.043           & 0.746          & 75.73           & 59.09          & 66.38          & 79.44          & 63.87         & 70.81  & 80.50 & 66.54 & 72.86          \\ 
MVSNet\cite{mvsnet}     & 0.444         & 0.741 & 0.592 & 82.93          & 62.71 & 71.42 & 88.58          & 68.70 & 77.38 & 89.85 & 70.11 & 78.76\\

\specialrule{.16em}{.08em}{.081em}
\end{tabular}%
\label{table:dtu}
\end{table*}

\begin{figure*}[!t]
    \centering
    \includegraphics[width=\linewidth]{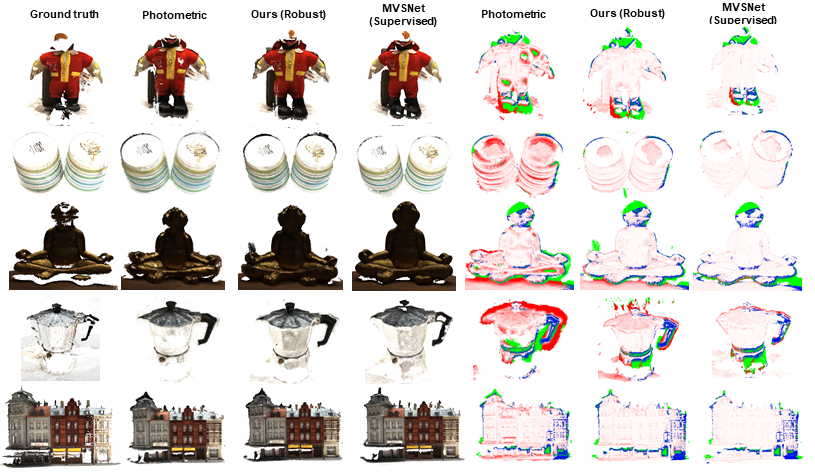}
    \caption{Left to right: a) Ground truth 3D scan, b) result with baseline photo-loss, c) result with robust photo-loss, d) result using a supervised approach (MVSNet~\cite{mvsnet}), e-g) corresponding error maps. For the error maps, points marked blue/green are masked out in evaluation. Magnitude of error is represented by variation from white-red in increasing order. Note how our method reconstructs areas not captured by the ground truth scan- doors and walls for the building in last row, complete face of statue in third row. Best viewed in color.}
    \label{fig:results}
\end{figure*} 

\section{Experiments}

We now describe the evaluation of our proposed models. The primary dataset of evaluation is the DTU MVS dataset \cite{jensen2014large}. In section \ref{dtu_setup} we describe the DTU dataset and our training and evaluation setup, and discuss our results, qualitatively and quantatively. Next, we perform rigorous ablation studies on the effects of various components of the robust loss function we propose (Section \ref{ablation}). We also show in Section \ref{eth_exp} that our method can allow us to adapt pretrained models to datasets without using ground-truth, by finetuning using our robust photometric consistency loss. Lastly, we demonstrate the generalization of our model to another dataset without finetuning (Section \ref{tnt_exp}).

\subsection{Benchmarking on DTU} \label{dtu_setup}
The DTU MVS dataset contains scans of 124 different scenes with 3D structure and high-resolution RGB images captured using a robotic arm. For each scene, there are 49 images whose camera poses are known with high accuracy. We use the same train-val-test split as used in SurfaceNet ~\cite{ji2017surfacenet} and MVSNet~\cite{mvsnet}.

As in MVSNet, for a given reference image of one scan, its neighboring images for input to the network (N) are selected using a view-selection score \cite{zhang2015joint}, which uses the sparse point cloud and camera baselines to pick the most suitable neighboring views for a given reference view. We similarly use neighboring $M$ views during training for self-supervision with the top-$K$ loss.

\subsubsection{Training setup} \label{tsetup}
For training, we scale the DTU images to 640x512 resolution. All of our networks are trained with $N=3$, such that during each iteration, one reference view and 2 novel views are used for predicting a depth map. For our top-$K$ aggregation based robust photometric loss, we use $M=6$ and $K=3$. Thus, 6 neighboring views are used to calculate the photometric loss volume, and per pixel the best 3 are selected. We later discuss the effect of varying $K$.

For evaluation on the test set, depth maps are generated at image resolution 640x512. The $d_{min}$ and $d_{max}$ for the plane sweep volume generation in the network is set to 425mm and 935mm respectively.

\begin{figure*}[ht]
  \minipage{0.32\textwidth}
    \includegraphics[width=\linewidth]{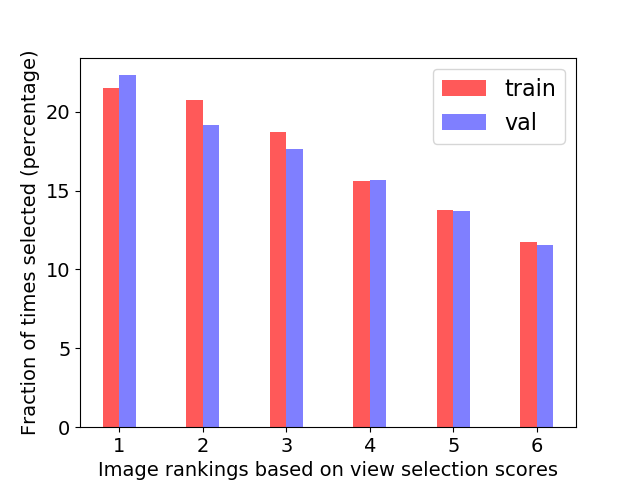}
    \caption{Frequency with which pixels from differently ranked images are picked as valid contributors to the top-$K$ photo-loss. The input images are ranked based on the view selection scores as detailed in Section~\ref{tsetup}.}
    \label{fig:topk_counts}
  \endminipage
  \hfill
  \minipage{0.32\textwidth}
    \includegraphics[width=\linewidth]{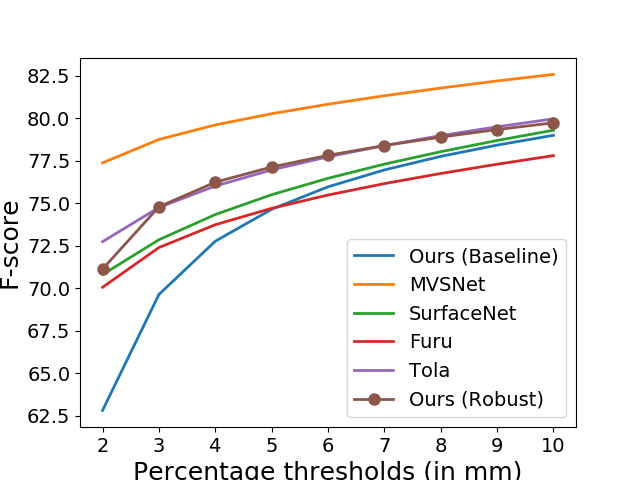}
    \caption{Comparison of different models on the \textit{DTU}'s evaluation set~\cite{aanaes2016large} using the F-score metric proposed in~\cite{knapitsch2017tanks}. We see that our model trained with robust loss consistently outperforms the baseline and several classical methods.}
    \label{fig:fscores}
  \endminipage
  \hfill
  \minipage{0.31\textwidth}
    \includegraphics[width=\linewidth]{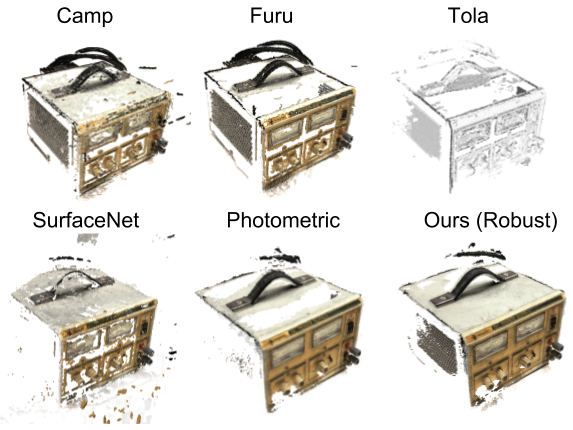}
    \caption{An example of how our proposed technique improves completeness over other methods in low-texture regions. Our result is a smooth dense reconstruction with significantly fewer holes or missing regions.}
    \label{fig:texture}
  \endminipage
\end{figure*}

\subsubsection{Results on DTU}
 
We evaluate our models on the test split of the DTU dataset\cite{jensen2014large} using the officially prescribed metrics:
 \begin{itemize}[noitemsep,topsep=0pt]
     \item \textit{Accuracy} : The mean distance of the reconstructed points from the ground truth.
     \item \textit{Completion} : The mean distance of the ground truth points to the reconstruction.
     \item \textit{Overall} : the mean of accuracy and completion.
 \end{itemize}
 Additionally, we report the percentage metric and f-score (which measures the overall accuracy and completeness of the point cloud) as used in the Tanks and Temples benchmark \cite{knapitsch2017tanks}.

We quantitatively evaluate three unsupervised models, namely: 
\begin{itemize}[noitemsep,topsep=0pt]
    \item  \emph{Photometric}: This model uses a combination of the naive photometric image reconstruction loss as in Equation~\ref{naive}, along with SSIM and Smooth loss.
    \item  \emph{Photometric + first order loss}: We replace the naive photometric loss with our proposed first order gradient consistency loss of Equation~\ref{gipuma}, which makes the network much more robust to local lighting variations (denoted as Photometric + G in Table \ref{table:dtu}).
    \item \emph{Robust:} Our best model, which combines both the first order gradient consistency loss and the top-$K$ view aggregation scheme.
\end{itemize}

To place our results in context, in addition to the unsupervised photometric setting, we compare our models against two classical methods (Furukawa et. al. and Tola et al.) \cite{furukawa2010accurate, Tola2011EfficientLM} , and two more recent deep learning methods \emph{that are fully-supervised}, SurfaceNet \cite{ji2017surfacenet} and MVSNet \cite{mvsnet}. To the best of our knowledge, we are not aware of any other existing deep-learning based models that learn this task in an unsupervised manner.

We find that for our model, the one with the robust loss, significantly outperforms the variants without it across all metrics. In order to characterize the performance of our model, we further compute the percentage metrics for distance thresholds up to $10mm$ and report the f-score plot in Figure~\ref{fig:fscores}. As reported in Table~\ref{table:dtu}, while our model struggles at a high resolution ($<1mm$), we outperform all other methods (except the fully-supervised MVSNet model) on increasing resolutions. This indicates that while some classical methods are more accurate compared to ours in very low thresholds, our approach produces fewer outliers. The quantitative results from Table~\ref{table:dtu} and qualitative visualizations of the errors in Figure~\ref{fig:results} show that our robust model leads to higher quality reconstructions. Figure~\ref{fig:texture} shows superior performance of our model in low-texture regions.

\subsection{Ablation studies}\label{ablation}
This section analyzes the influence of several design choices involved in our system, and further highlights the importance of the robust losses in our training setup.

\vspace{1mm}
\noindent \textbf{Top-$K$ Selection Frequency.} In order to characterize the top-$K$ choice selection, we visualize the frequency with which pixels from different views are selected for photo-consistency. We run the trained model on the training and validation datasets for 50 iterations and store frequency counts of top-$K$ operations which are shown in Figure~\ref{fig:topk_counts}. 
We can observe two things: 1) A view's selection frequency is directly proportional to its view selection score. This validates that the view-selection criterion used for picking image sets for training corresponds directly to photo-consistency, 2) More than $50\%$ of selections are from views ranked lower than 2 which explains why adding the flexibility of accumulating evidence from additional images leads to better performance.




\vspace{1mm}
\noindent \textbf{Top-$K$ Selection Threshold.} We ablate the effect of varying $K$ in our robust loss formulation. As can be seen from Table~\ref{tab:vary_k}, using $K = 3$ i.e. $50\%$ of the non-reference images has a substantially better validation accuracy. Note that for validation, we use accuracies against the ground truth depth maps. We report percentages of pixels where the absolute difference in depth values is under $3\%$. 

\begin{table}[ht]
    \centering
    \caption{Ablation study of models with various combinations of loss functions, in terms of validation accuracy against ground truth depth maps of DTU MVS datasets. B signifies the naive baseline photometric loss, as given in eqn. \ref{naive}. G is the first order gradient consistent loss. Our robust model is a combination of G, SSIM, Smooth and top-$K$ aggregation. }
    \begin{small}

    \begin{tabular}{lccc} 
        \toprule
        Loss used & L1 error &  $\% < 1mm$ & $\% < 3mm$ \\
        \midrule
        B + SSIM & 6.57 & 30.93 & 55.07\\
        \midrule
     B + Smooth & 5.92 & 42.52 & 63.05\\
        \midrule
         B + Smooth \\ + SSIM (our baseline)& 4.98 & 49.37 & 72.92\\
        \midrule
         G + Smooth + SSIM  & 5.33 & 61.92 & 77.29\\
          \midrule
         G+Smooth+SSIM \\ w/ top k (our robust) & \textbf{4.06} & \textbf{65.33} & \textbf{81.08}\\ 
        \bottomrule
    \end{tabular}
    
    \end{small}
    \label{tab:vary_loss}
\end{table}

\begin{table}[ht]
    \centering
    \caption{Performance comparison as the $K$ in our robust photo-consistency loss varies. Results for using best $25\%$, $50\%$ and $100\%$ of warping losses per-pixel.}
    \begin{small}
    \begin{tabular}{lccc}
        \toprule
        Method (M=6) & K=1 & K=3 & K=6 \\
        \midrule
        Validation Accuracy ($\%$) & 75.59 & \textbf{81.08} & 77.99\\
        \bottomrule
    \end{tabular}
    \end{small}
    \label{tab:vary_k}
\end{table}

\vspace{1mm}
\noindent \textbf{Impact of loss terms.}
We perform ablations to analyze the different components of our robust photometric loss. Although our models are trained in an  unsupervised manner, we use the ground truth depth maps of the validation set of DTU to evaluate their performances. In particular, we evaluate the methods on 3 metrics : 1) Absolute difference between predicted and ground truth depths (in $mm$, lower is better) 2) Percentage of predicted depths within $1mm$ of ground truth (higher is better) 3) Percentage of predicted depths within $3mm$ of ground truth. The detailed quantitative results are provided in Table \ref{tab:vary_loss}. We observe that both the proposed modifications over the naive baseline yield significant improvements.

\subsection{Fine-tuning on ETH3D} \label{eth_exp}
We also test the effectiveness of the robust consistency formulation as a means of fine-tuning pretrained models on unseen datasets without available annotations. 
On the low-res many view dataset of the ETH3D benchmark\cite{schoeps2017cvpr}, we compare results of a pretrained MVSNet model with one that is fine-tuned on the train split of ETH3D. 
For fusion of depth maps, we run Fusibile\cite{galliani2015massively} with identical hyperparameters for each. 
The results in Table~\ref{tab:eth} demonstrate that fine-tuning with our proposed loss, in the absence of available 3D ground truth annotations, improves performance. 
\begin{table}[ht]
    \centering
    \caption{Effect of fine-tuning on the low-res many view ETH3D dataset. All metrics are represented as ($\%$) and higher is better.}
    \begin{small}
    \begin{tabular}{lccc}
        \toprule
        Method & F1 score & Accuracy & Completeness \\
        \midrule
        Pretrained MVSNet & 16.91 & 17.51 & 19.59 \\
        Fine-tuned MVSNet & \textbf{17.31} & \textbf{18.31} & \textbf{19.68} \\
        \bottomrule
    \end{tabular}
    \end{small}
    \label{tab:eth}
\end{table}

 \subsection{Generalization on Tanks and Temples } \label{tnt_exp}
We perform an experiment to check the generalization ability of our network. Since the network has explicitly been trained to match image correspondences rather than memorize scene priors, our hypothesis is that it should generalize well to completely unseen data. We select the Tanks and Temples dataset for this purpose, which contains high-res images of outdoor scenes of large objects. We use our model trained on DTU on images from this dataset, without any fine-tuning. We downscale the images to 832x512 resolution and use 256 depth intervals for the plane-sweep volume. The results are visualized in Figure~ \ref{fig:tnt}. More extensive results are provided in the supplemental. However, we do note that the very high depth range of scenes in open-world datasets like Tanks and Temples are not amenable to the current deep architectures for MVS, as they all rely on some sort of volume formulation. Thus to sample depths at a finer resolution for higher quality reconstructions becomes extremely computationally expensive, and is perhaps a promising direction for future work.

\begin{figure}
    \centering
    \includegraphics[scale=0.25]{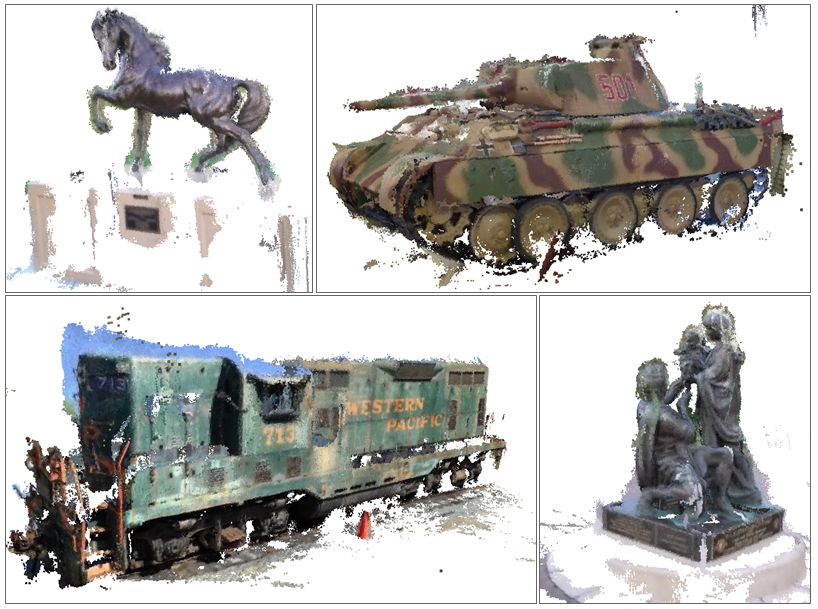}
    \caption{Generalization result of our robust model on the Tanks and Temples\cite{knapitsch2017tanks} dataset. Without any finetuning, our robust model provides reasonable reconstructions.}
    \label{fig:tnt}
\end{figure}

\section{Discussion}
We presented an unsupervised learning based approach for multi-view stereopsis, and proposed robust photometric losses to learn effectively in this setting. This is however, only an initial attempt, and further efforts are required to realize the potential of unsupervised methods for this task. 
We are however optimistic, as an unsupervised approach is more scalable as large amounts of training data can be more easily acquired.
In addition, as our experiments demonstrated, these unsupervised methods can be used in conjunction with, and further allow us to improve over supervised methods, thereby allowing us to leverage both, the benefits of supervision along with the scalability of unsupervised methods.
Lastly, we also hope that the proposed robust photometric loss formulation would be more broadly applicable for unsupervised 3D prediction approaches.



\section*{Acknowledgements}
This project is supported by Carnegie Mellon University's Mobility21 National University Transportation Center, which is sponsored by the US Department of Transportation.

\section*{Supplementary}

\subsection{Overview}
In this document we provide additional quantitative results, technical details and more qualitative examples of the results on both DTU\cite{jensen2014large} and Tanks and Temples dataset\cite{knapitsch2017tanks}.

In Section~\ref{model}, we present details of the model architecture used for all our experiments. Section~\ref{loss} describes the mathematical formulation of the various loss terms used. In Section~\ref{dtu_results} we show some elaborate quantitative results on the DTU dataset and in Section~\ref{more_results} we visualize the qualitative output results for the remaining instances in the DTU dataset. Higher resolution qualitative results on the Tanks and Temples dataset are included in Section~\ref{tnt}.

\subsection{Model Architecture}\label{model}
We adapt the same network architecture as MVSNet\cite{mvsnet} to emphasize that our key contribution is the loss objective used for training a standard network with sufficient capacity for this task.

Every input image is first passed through a feature extraction network having shared weights for all images. For this network, we use use an 8-layer CNN having batch-normalization and ReLU after every convolution operation till the penultimate layer. The last layer produces a 32 channel downsized feature map for every image. Using the differentiable homography formulation, the feature maps are warped into different fronto-parallel planes of the reference camera at 128 depth values to form one cost volume per non-reference image. All such cost volumes are aggregated into a single volume using a variance-based cost metric. Note that MVSNet uses 256 depth values for the cost volume during training. Since this setup does not fit in our 12GB GPU memory, we use only 128 depth values. This reduction does play a role in the output reconstruction quality, but we leave this optimization for future work since our contributions hold nonetheless.

In order to refine the cost volume and incorporate smooth variations of the depth values, we use a three layer 3D U-Net. An initial estimate of the predicted depth map can be obtained by performing a soft $argmin$ operation along the depth channel. Unlike the $winner-take-all$ approach which requires the non-differentiable $argmax$ operation, such a soft aggregation of volumes allows for sub-pixel accuracies while being amenable to training due to its differentiability. Thus, in spite of the discretization of depth value for constructing the cost volume, the resulting depth map follows a continuous distribution.

The resulting probability distribution which the output volume represents is likely to be containing outliers and would not necessarily contain a single peak. To account for this, a notion of depth estimate quality is established wherein the quality of estimate at any pixel is defined to be the sum of the probabilities over the four nearest depth hypotheses. This estimate is then filtered at a threshold of $0.8$ and applied as a mask to the output volume. The predicted depth map is then concatenated to the reference image and passed through a four-layer CNN to output a depth residual map. The final depth map is obtained by adding the residual map to the initial estimated depth map. For complete details of the hyperparameters, we refer the reader to the MVSNet\cite{mvsnet} paper and it's corresponding supplemental.

\subsection{Loss functions}\label{loss}
While minimizing the photometric consistency loss obtained by view synthesis is our primary objective, we make use of two additional ingredients in the loss objective to improve model performance.
We augment our robust photometric loss with two additional losses, namely image-patch level structured similarity loss and an image-aware smoothness loss on the depth map's gradients.\\

\textbf{SSIM:} We take cues from recent works\cite{Zhou2017UnsupervisedLO, godard2017unsupervised, mahjourian2018unsupervised} showing the effectiveness of perceptual losses for evaluating the quality of image predictions. Similarly, we also use the structured similarity (SSIM) as  a loss term for training.
The SSIM similarity betwen two image patches is given by :
\begin{equation}
    SSIM(x,y) = \frac{(2\mu_x\mu_y + c_1)(2\sigma_{xy} + c_2)}{(\mu_x^2 + \mu_y^2 +c_1)(\sigma_x +  \sigma_y + c_2)}
\end{equation}
Here, $\mu$ and $\sigma$ are the local mean and variance respectively. We compute $\mu_x$, $\mu_y$, $\sigma_x$, $\sigma_y$ using average pooling and set $c_1 = 0.01^2$ and $c_2 = 0.03^2$. Since higher values of SSIM are desirable, we minimize its distance to 1 which is the highest attainable similarity value.
The SSIM loss for an image pair then becomes :
\begin{equation}
    L_{SSIM} = \sum_{ij} \left[ 1 - SSIM(I_s^{ij} , \hat{I_s^{ij}}) \right] M_s^{ij}
\end{equation}
Here, $M_s^{ij}$ is a mask which excludes all pixels whose projections after inverse warping lie outside the source image. As observed in \cite{mahjourian2018unsupervised}, ignoring such regions improves depth predictions around the boundaries.
We apply the SSIM loss only between the reference image and two nearest images ranked by view selection score.

\textbf{Depth Smoothness loss:} In order to encourage smoother gradient changes and allow sharp depth discontinues at pixels corresponding to sharp changes in the image, it is important to regularize the depth estimates. Similar to \cite{mahjourian2018unsupervised}, we add an $l_1$ penalty on the depth gradients.

\begin{equation}
    L_{Smooth} = \sum_{ij} ||\nabla_xD^{ij}||e^{-||\nabla_xI^{ij}||} +\\ ||\nabla_yD^{ij}||e^{-||\nabla_yI^{ij}||}
\end{equation}

\subsection{Quantitative Results}\label{dtu_results}
The DTU dataset's evaluation script measures the reconstruction quality in terms of accuracy and completeness while also reporting their median values and variances. We list these results for two classical (top), two supervised learning based (bottom), and three unsupervised learning (middle) methods in Table~\ref{tab:results}. As noted by \cite{mvsnet}, SurfaceNet\cite{ji2017surfacenet} used their own script for evaluation. However, we use the released DTU evaluation benchmark scheme for reporting results from all the methods.

\begin{figure*}[!t]
\centering
\begin{subfigure}{.45\textwidth}
  \centering
  \includegraphics[width=\linewidth]{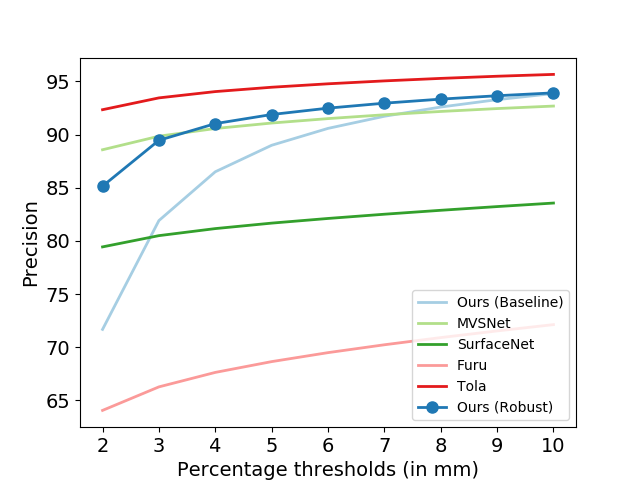}
  \label{fig:sub1}
\end{subfigure}%
\begin{subfigure}{.45\textwidth}
  \centering
  \includegraphics[width=\linewidth]{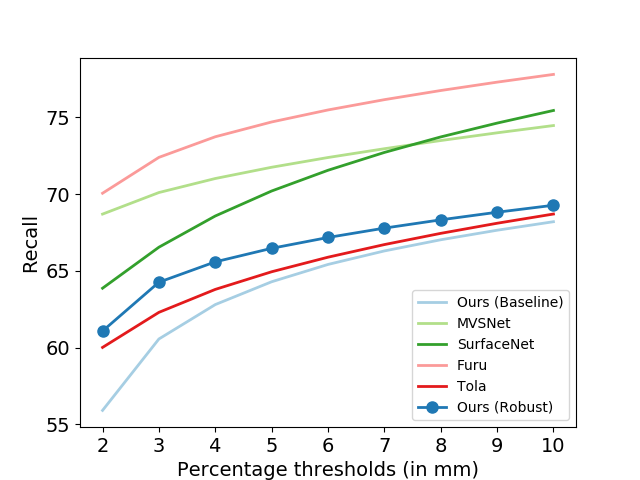}
  \label{fig:sub2}
\end{subfigure}
\caption{Percentage metrics on the DTU datasets as proposed in \cite{knapitsch2017tanks}. The f-scores reported in the main paper are computed using precision and recall which are displayed here. }
\label{fig:prec_recall}
\end{figure*}

Additionally, we show the comparison of two components of the percentage metric, precision and recall, that make up the f-score reported in the main paper, in Figure~\ref{fig:prec_recall}.

\begin{table*}[!h]
\centering
\caption{Quantitative results on the \textit{DTU}'s evaluation set \cite{aanaes2016large}. We evaluate two classical MVS methods (top), two learning based MVS methods (middle) and three variants of our unsupervised method using the distance metrics. For all columns, lower is better.}
\begin{tabular}{c c c c| c c c | c}
\specialrule{.16em}{.08em}{.08em}
           & \multicolumn{3}{c|}{Accuracy} & \multicolumn{3}{c|}{Completeness} & \multicolumn{1}{c}{Overall}\\
           & \multicolumn{3}{c|}{Mean Median Variance} & \multicolumn{3}{c|}{Mean Median Variance}\\ \hline

Furu \cite{furukawa2010accurate}    & 0.612 & 0.324 & 1.249 & 0.939 & 0.463 & 3.392 & 0.775 \\
Tola \cite{Tola2011EfficientLM}  & 0.343 & 0.210 & 0.385 & 1.190 & 0.492 & 5.319 & 0.766 \\
\hline
Ours (Baseline)    &  1.565 & 1.041 & 3.683 & 1.378 & 0.694 & 4.964 & 1.472\\
Ours (Baseline+G)    &  1.069 & 0.759 & 1.883 & 1.020 & 0.595 & 2.779 & 1.045\\
Ours (Robust)    &  0.881 & 0.673 & 1.075 & 1.073 & 0.617 & 3.418 & 0.977 \\
\hline
SurfaceNet\cite{ji2017surfacenet} & 0.450 & 0.254 & 1.270 & 1.043 & 0.285 & 5.594 & 0.746 \\
MVSNet\cite{mvsnet} & 0.444 & 0.307 & 0.436 & 0.741 & 0.399 & 2.501 & 0.592  \\

\specialrule{.16em}{.08em}{.08em}
\end{tabular}%
\label{tab:results}
\end{table*}

\subsection{Qualitative Results}\label{more_results}
\begin{figure*}[!t]
    \centering
    \includegraphics[width=0.98\linewidth]{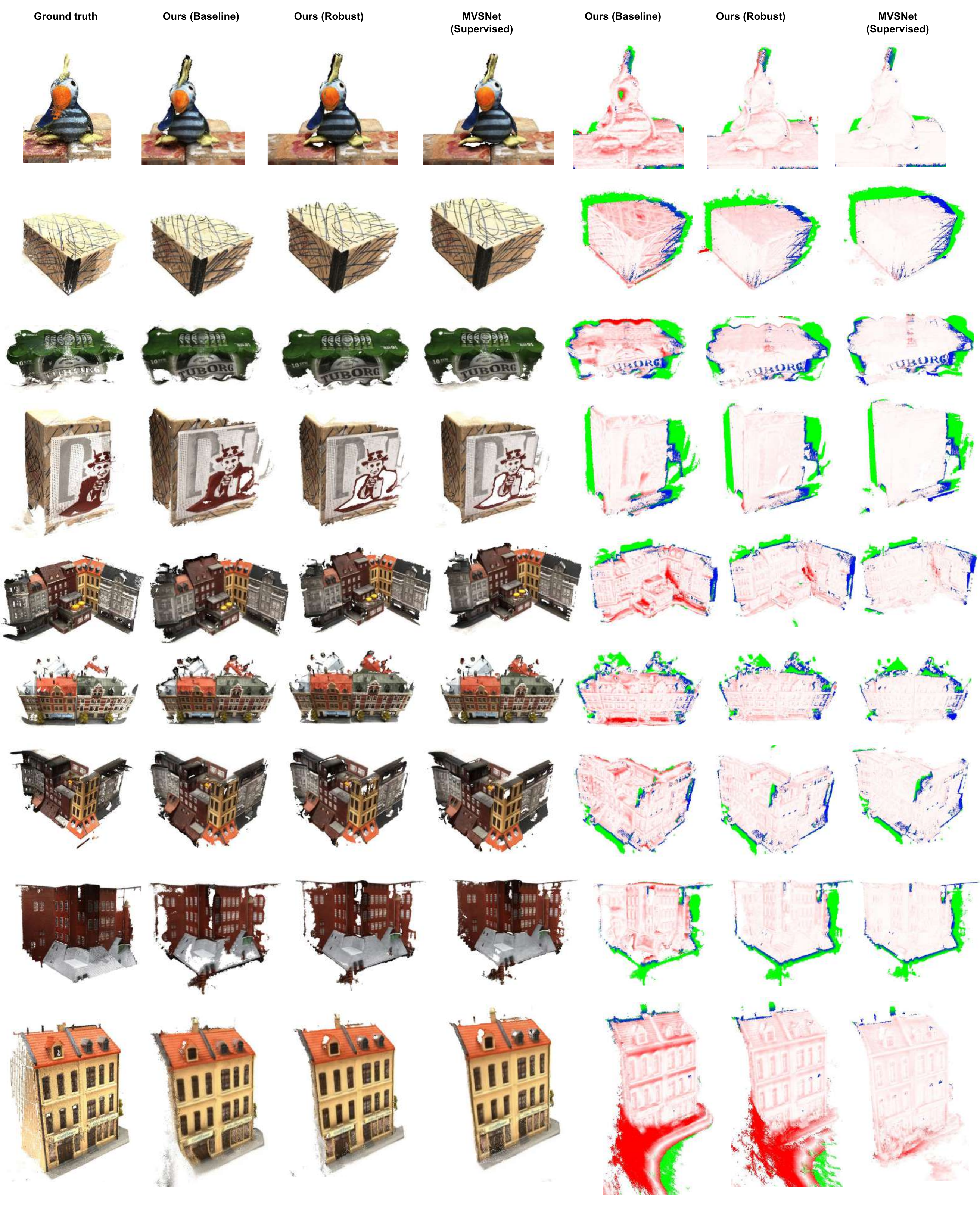}
    \caption{Predictions for the remaining instances of the DTU test set. Left to right: a) Ground truth 3D scan, b) result with baseline photo-loss, c) result with robust photo-loss, d) result using a supervised approach (MVSNet~\cite{mvsnet}), e-g) corresponding error maps. For the error maps, points marked blue/green are masked out in evaluation. Magnitude of error is represented by variation from white-red in increasing order. Best viewed in color.}
    \label{fig:results1}
\end{figure*}

\begin{figure*}[!t]
    \centering
    \includegraphics[width=0.98\linewidth]{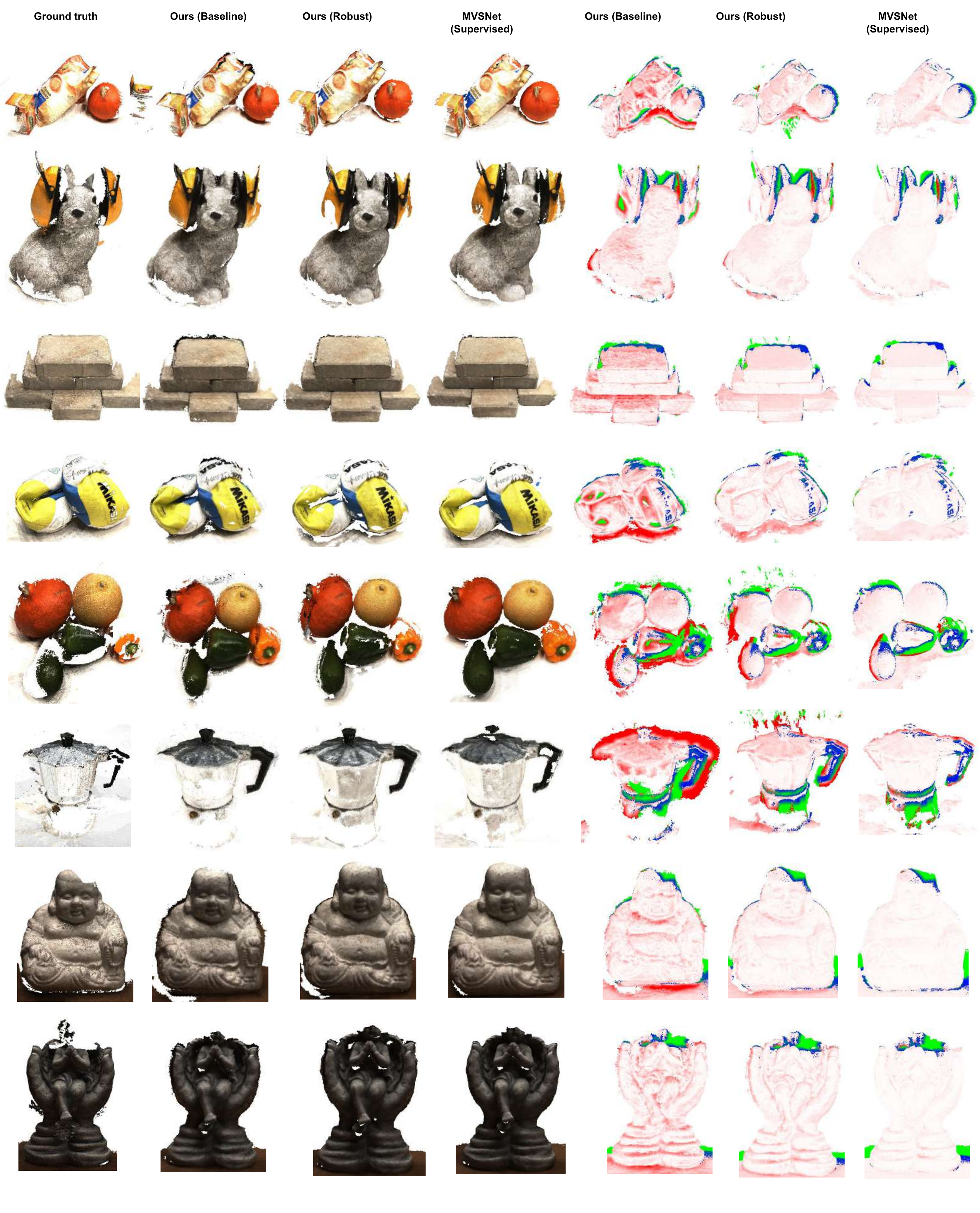}
    \caption{Predictions for the remaining instances of the DTU test set. Left to right: a) Ground truth 3D scan, b) result with baseline photo-loss, c) result with robust photo-loss, d) result using a supervised approach (MVSNet~\cite{mvsnet}), e-g) corresponding error maps. For the error maps, points marked blue/green are masked out in evaluation. Magnitude of error is represented by variation from white-red in increasing order. Best viewed in color.}
    \label{fig:results2}
\end{figure*}

Qualitative results for the remaining 17 instances of the DTU test set are presented in Figures~\ref{fig:results1},~\ref{fig:results2}.

\subsection{Tanks and Temples}\label{tnt}
\begin{figure*}[!t]
    \centering
    \includegraphics[width=0.98\linewidth]{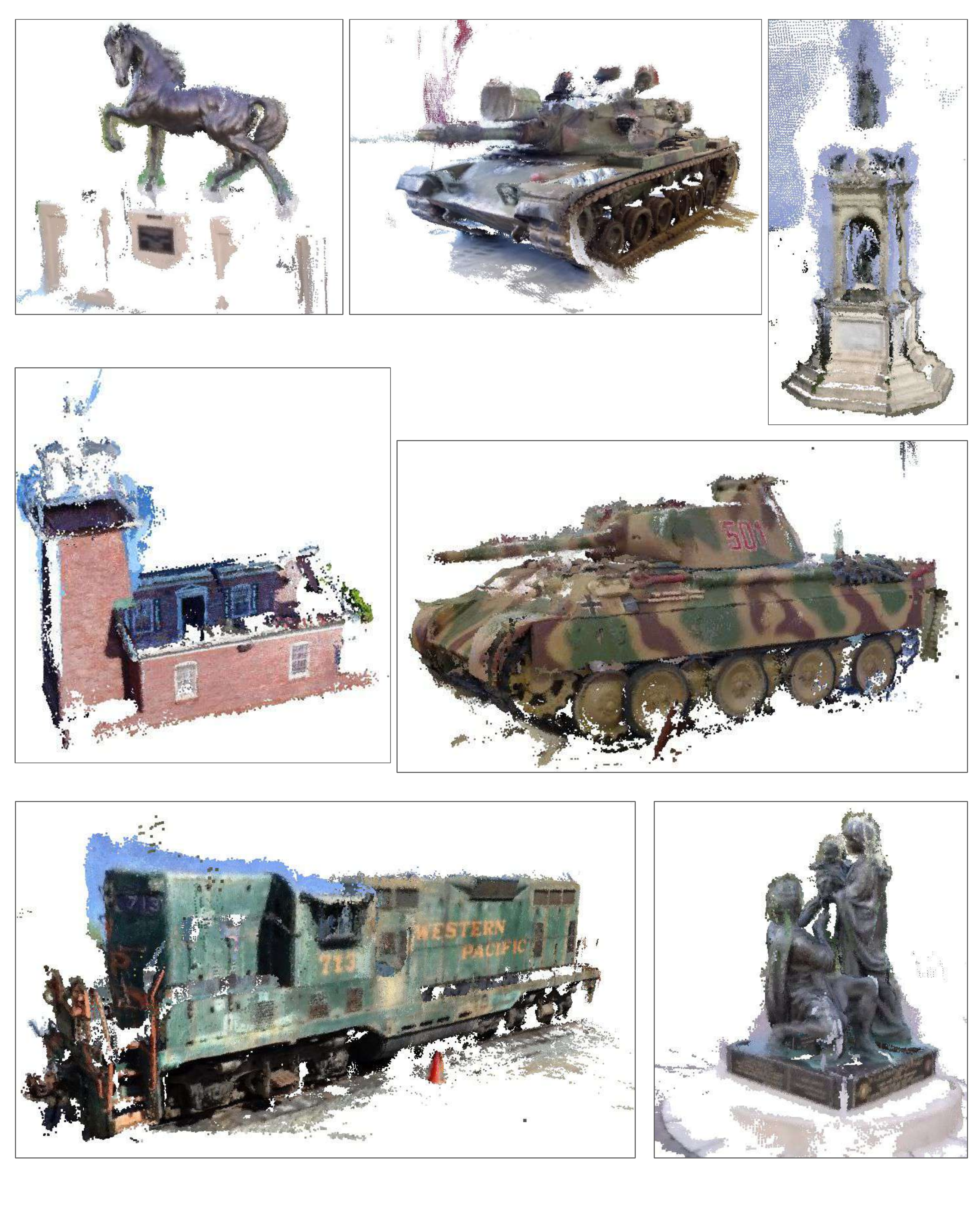}
    \caption{Generalization results of our robust model on the Tanks and Temples dataset without any finetuning. The results shown are from the Intermediate level instances of the dataset. The scene names from top left to bottom right: Horse, M60, Francis, Lighthouse, Panther, Train, Family. Best viewed in color.}
    \label{fig:results3}
\end{figure*}

We show qualitative results of our robust models on scenes from the Intermediate set of the Tanks and Temples dataset in Figure:~\ref{fig:results3}. The model has not been finetuned on the dataset but produces reasonable reconstructions demonstrating the learned photo-consistency behavior.

{\small
\bibliographystyle{ieee}
\bibliography{references}

\begin{thebibliography}{10}\itemsep=-1pt

\bibitem{aanaes2016large}
H.~Aan{\ae}s, R.~R. Jensen, G.~Vogiatzis, E.~Tola, and A.~B. Dahl.
\newblock Large-scale data for multiple-view stereopsis.
\newblock {\em International Journal of Computer Vision}, 120(2):153--168,
  2016.

\bibitem{abadi2016tensorflow}
M.~Abadi, P.~Barham, J.~Chen, Z.~Chen, A.~Davis, J.~Dean, M.~Devin,
  S.~Ghemawat, G.~Irving, M.~Isard, et~al.
\newblock Tensorflow: a system for large-scale machine learning.

\bibitem{barnes2009patchmatch}
C.~Barnes, E.~Shechtman, A.~Finkelstein, and D.~B. Goldman.
\newblock Patchmatch: A randomized correspondence algorithm for structural
  image editing.
\newblock {\em ACM Transactions on Graphics (ToG)}, 28(3):24, 2009.

\bibitem{furukawa2015multi}
Y.~Furukawa, C.~Hern{\'a}ndez, et~al.
\newblock Multi-view stereo: A tutorial.
\newblock {\em Foundations and Trends{\textregistered} in Computer Graphics and
  Vision}, 9(1-2):1--148, 2015.

\bibitem{furukawa2010accurate}
Y.~Furukawa and J.~Ponce.
\newblock Accurate, dense, and robust multiview stereopsis.
\newblock {\em IEEE transactions on pattern analysis and machine intelligence},
  32(8):1362--1376, 2010.

\bibitem{galliani2015massively}
S.~Galliani, K.~Lasinger, and K.~Schindler.
\newblock Massively parallel multiview stereopsis by surface normal diffusion.
\newblock In {\em Proceedings of the IEEE International Conference on Computer
  Vision}, pages 873--881, 2015.

\bibitem{Garg2016UnsupervisedCF}
R.~Garg, G.~VijayKumarB., and I.~D. Reid.
\newblock Unsupervised cnn for single view depth estimation: Geometry to the
  rescue.
\newblock In {\em ECCV}, 2016.

\bibitem{Geiger2012AreWR}
A.~Geiger, P.~Lenz, and R.~Urtasun.
\newblock Are we ready for autonomous driving? the kitti vision benchmark
  suite.
\newblock {\em 2012 IEEE Conference on Computer Vision and Pattern
  Recognition}, pages 3354--3361, 2012.

\bibitem{godard2017unsupervised}
C.~Godard, O.~Mac~Aodha, and G.~J. Brostow.
\newblock Unsupervised monocular depth estimation with left-right consistency.
\newblock In {\em CVPR}, volume~2, page~7, 2017.

\bibitem{goesele2007multi}
M.~Goesele, N.~Snavely, B.~Curless, H.~Hoppe, and S.~M. Seitz.
\newblock Multi-view stereo for community photo collections.
\newblock 2007.

\bibitem{Han2015MatchNetUF}
X.~Han, T.~Leung, Y.~Jia, R.~Sukthankar, and A.~C. Berg.
\newblock Matchnet: Unifying feature and metric learning for patch-based
  matching.
\newblock {\em 2015 IEEE Conference on Computer Vision and Pattern Recognition
  (CVPR)}, pages 3279--3286, 2015.

\bibitem{hartmann2017learned}
W.~Hartmann, S.~Galliani, M.~Havlena, L.~Van~Gool, and K.~Schindler.
\newblock Learned multi-patch similarity.
\newblock In {\em 2017 IEEE International Conference on Computer Vision
  (ICCV)}, pages 1595--1603. IEEE, 2017.

\bibitem{hirschmuller2008stereo}
H.~Hirschmuller.
\newblock Stereo processing by semiglobal matching and mutual information.
\newblock {\em IEEE Transactions on pattern analysis and machine intelligence},
  30(2):328--341, 2008.

\bibitem{hu2012least}
X.~Hu and P.~Mordohai.
\newblock Least commitment, viewpoint-based, multi-view stereo.
\newblock In {\em 3D Imaging, Modeling, Processing, Visualization and
  Transmission (3DIMPVT), 2012 Second International Conference on}, pages
  531--538. IEEE, 2012.

\bibitem{huang2018deepmvs}
P.-H. Huang, K.~Matzen, J.~Kopf, N.~Ahuja, and J.-B. Huang.
\newblock Deepmvs: Learning multi-view stereopsis.
\newblock In {\em Proceedings of the IEEE Conference on Computer Vision and
  Pattern Recognition}, pages 2821--2830, 2018.

\bibitem{DBLP:journals/corr/JaderbergSZK15}
M.~Jaderberg, K.~Simonyan, A.~Zisserman, and K.~Kavukcuoglu.
\newblock Spatial transformer networks.
\newblock {\em CoRR}, abs/1506.02025, 2015.

\bibitem{jancosek2011multi}
M.~Jancosek and T.~Pajdla.
\newblock Multi-view reconstruction preserving weakly-supported surfaces.
\newblock In {\em Computer Vision and Pattern Recognition (CVPR), 2011 IEEE
  Conference on}, pages 3121--3128. IEEE, 2011.

\bibitem{jensen2014large}
R.~Jensen, A.~Dahl, G.~Vogiatzis, E.~Tola, and H.~Aan{\ae}s.
\newblock Large scale multi-view stereopsis evaluation.
\newblock In {\em 2014 IEEE Conference on Computer Vision and Pattern
  Recognition}, pages 406--413. IEEE, 2014.

\bibitem{ji2017surfacenet}
M.~Ji, J.~Gall, H.~Zheng, Y.~Liu, and L.~Fang.
\newblock Surfacenet: An end-to-end 3d neural network for multiview stereopsis.
\newblock In {\em Proceedings of the IEEE International Conference on Computer
  Vision}, pages 2307--2315, 2017.

\bibitem{kang2001handling}
S.~B. Kang, R.~Szeliski, and J.~Chai.
\newblock Handling occlusions in dense multi-view stereo.
\newblock In {\em Computer Vision and Pattern Recognition, 2001. CVPR 2001.
  Proceedings of the 2001 IEEE Computer Society Conference on}, volume~1, pages
  I--I. IEEE, 2001.

\bibitem{kar2017learning}
A.~Kar, C.~H{\"a}ne, and J.~Malik.
\newblock Learning a multi-view stereo machine.
\newblock In {\em Advances in neural information processing systems}, pages
  365--376, 2017.

\bibitem{kendall2017end}
A.~Kendall, H.~Martirosyan, S.~Dasgupta, P.~Henry, R.~Kennedy, A.~Bachrach, and
  A.~Bry.
\newblock End-to-end learning of geometry and context for deep stereo
  regression.
\newblock {\em CoRR, vol. abs/1703.04309}, 2017.

\bibitem{Kingma2014AdamAM}
D.~P. Kingma and J.~Ba.
\newblock Adam: A method for stochastic optimization.
\newblock {\em CoRR}, abs/1412.6980, 2014.

\bibitem{DBLP:conf/eccv/KlodtV18}
M.~Klodt and A.~Vedaldi.
\newblock Supervising the new with the old: Learning {SFM} from {SFM}.
\newblock In {\em {ECCV} {(10)}}, volume 11214 of {\em Lecture Notes in
  Computer Science}, pages 713--728. Springer, 2018.

\bibitem{knapitsch2017tanks}
A.~Knapitsch, J.~Park, Q.-Y. Zhou, and V.~Koltun.
\newblock Tanks and temples: Benchmarking large-scale scene reconstruction.
\newblock {\em ACM Transactions on Graphics (ToG)}, 36(4):78, 2017.

\bibitem{Kuznietsov2017SemiSupervisedDL}
Y.~Kuznietsov, J.~St{\"u}ckler, and B.~Leibe.
\newblock Semi-supervised deep learning for monocular depth map prediction.
\newblock {\em 2017 IEEE Conference on Computer Vision and Pattern Recognition
  (CVPR)}, pages 2215--2223, 2017.

\bibitem{mahjourian2018unsupervised}
R.~Mahjourian, M.~Wicke, and A.~Angelova.
\newblock Unsupervised learning of depth and ego-motion from monocular video
  using 3d geometric constraints.
\newblock In {\em Proceedings of the IEEE Conference on Computer Vision and
  Pattern Recognition}, pages 5667--5675, 2018.

\bibitem{schonberger2016pixelwise}
J.~L. Sch{\"o}nberger, E.~Zheng, J.-M. Frahm, and M.~Pollefeys.
\newblock Pixelwise view selection for unstructured multi-view stereo.
\newblock In {\em European Conference on Computer Vision}, pages 501--518.
  Springer, 2016.

\bibitem{schops2017multi}
T.~Sch{\"o}ps, J.~L. Sch{\"o}nberger, S.~Galliani, T.~Sattler, K.~Schindler,
  M.~Pollefeys, and A.~Geiger.
\newblock A multi-view stereo benchmark with high-resolution images and
  multi-camera videos.
\newblock In {\em Conference on Computer Vision and Pattern Recognition
  (CVPR)}, volume 2017, 2017.

\bibitem{schoeps2017cvpr}
T.~Sch\"ops, J.~L. Sch\"onberger, S.~Galliani, T.~Sattler, K.~Schindler,
  M.~Pollefeys, and A.~Geiger.
\newblock A multi-view stereo benchmark with high-resolution images and
  multi-camera videos.
\newblock In {\em Conference on Computer Vision and Pattern Recognition
  (CVPR)}, 2017.

\bibitem{seitz2006comparison}
S.~M. Seitz, B.~Curless, J.~Diebel, D.~Scharstein, and R.~Szeliski.
\newblock A comparison and evaluation of multi-view stereo reconstruction
  algorithms.
\newblock In {\em 2006 IEEE Computer Society Conference on Computer Vision and
  Pattern Recognition (CVPR'06)}, volume~1, pages 519--528. IEEE, 2006.

\bibitem{shen2013accurate}
S.~Shen.
\newblock Accurate multiple view 3d reconstruction using patch-based stereo for
  large-scale scenes.
\newblock {\em IEEE transactions on image processing}, 22(5):1901--1914, 2013.

\bibitem{Tola2011EfficientLM}
E.~Tola, C.~Strecha, and P.~Fua.
\newblock Efficient large-scale multi-view stereo for ultra high-resolution
  image sets.
\newblock {\em Machine Vision and Applications}, 23:903--920, 2011.

\bibitem{wang2018mvdepthnet}
K.~Wang and S.~Shen.
\newblock Mvdepthnet: Real-time multiview depth estimation neural network.
\newblock In {\em 2018 International Conference on 3D Vision (3DV)}, pages
  248--257. IEEE, 2018.

\bibitem{Xie2016Deep3DFA}
J.~Xie, R.~B. Girshick, and A.~Farhadi.
\newblock Deep3d: Fully automatic 2d-to-3d video conversion with deep
  convolutional neural networks.
\newblock In {\em ECCV}, 2016.

\bibitem{mvsnet}
Y.~Yao, Z.~Luo, S.~Li, T.~Fang, and L.~Quan.
\newblock Mvsnet: Depth inference for unstructured multi-view stereo.
\newblock In {\em Proceedings of the European Conference on Computer Vision
  (ECCV)}, pages 767--783, 2018.

\bibitem{zach2008fast}
C.~Zach.
\newblock Fast and high quality fusion of depth maps.
\newblock In {\em Proceedings of the international symposium on 3D data
  processing, visualization and transmission (3DPVT)}, volume~1. Citeseer,
  2008.

\bibitem{zbontar2016stereo}
J.~Zbontar and Y.~LeCun.
\newblock Stereo matching by training a convolutional neural network to compare
  image patches.
\newblock {\em Journal of Machine Learning Research}, 17(1-32):2, 2016.

\bibitem{zhang2015joint}
R.~Zhang, S.~Li, T.~Fang, S.~Zhu, and L.~Quan.
\newblock Joint camera clustering and surface segmentation for large-scale
  multi-view stereo.
\newblock In {\em Proceedings of the IEEE International Conference on Computer
  Vision}, pages 2084--2092, 2015.

\bibitem{Zhang2018ActiveStereoNetES}
Y.~Zhang, S.~Khamis, C.~Rhemann, J.~P.~C. Valentin, A.~Kowdle, V.~Tankovich,
  M.~Schoenberg, S.~Izadi, T.~A. Funkhouser, and S.~R. Fanello.
\newblock Activestereonet: End-to-end self-supervised learning for active
  stereo systems.
\newblock {\em CoRR}, abs/1807.06009, 2018.

\bibitem{zheng2014patchmatch}
E.~Zheng, E.~Dunn, V.~Jojic, and J.-M. Frahm.
\newblock Patchmatch based joint view selection and depthmap estimation.
\newblock In {\em Proceedings of the IEEE Conference on Computer Vision and
  Pattern Recognition}, pages 1510--1517, 2014.

\bibitem{Zhong2017SelfSupervisedLF}
Y.~Zhong, Y.~Dai, and H.~Li.
\newblock Self-supervised learning for stereo matching with self-improving
  ability.
\newblock {\em CoRR}, abs/1709.00930, 2017.

\bibitem{Zhou2017UnsupervisedLO}
T.~Zhou, M.~Brown, N.~Snavely, and D.~G. Lowe.
\newblock Unsupervised learning of depth and ego-motion from video.
\newblock {\em 2017 IEEE Conference on Computer Vision and Pattern Recognition
  (CVPR)}, pages 6612--6619, 2017.

\end{thebibliography}
}

\end{document}